\documentclass{article}

\usepackage[final, nonatbib]{neurips_2023}

\usepackage[utf8]{inputenc}               % allow utf-8 input
\usepackage[T1]{fontenc}                  % use 8-bit T1 fonts
\usepackage{hyperref}                     % hyperlinks
\usepackage{url}                          % simple URL typesetting
\usepackage{booktabs}                     % professional-quality tables
\usepackage{amsfonts}                     % blackboard math symbols
\usepackage{nicefrac}                     % compact symbols for 1/2, etc.
\usepackage{microtype}                    % microtypography
\usepackage{graphicx}                     % figure 
\usepackage{subcaption}                   % caption in subfigure 
\usepackage{amsmath}                      % equation 
\usepackage{multirow}                     % multiple row for table

\usepackage{MnSymbol}                     % star symbol
\usepackage{xcolor}                       % text color 
\usepackage{boldline}                     % line width in table 

\usepackage{caption}
\usepackage{subcaption}
\usepackage{floatrow}
\newfloatcommand{capbtabbox}{table}[][\FBwidth]

\newcommand\R{\mathbb{R}}                 % real number 
\usepackage{bbm}                          % indicator function 

\usepackage{algorithm2e}
\RestyleAlgo{ruled} 
\SetKwComment{Comment}{/* }{ */}
\SetKwInput{kwInit}{Init}

\hypersetup{
    colorlinks=true,
    linkcolor=magenta,
    filecolor=magenta,      
    urlcolor=magenta,
    pdftitle={Overleaf Example},
    pdfpagemode=FullScreen,
    }

\title{Tree-Regularized Tabular Embeddings}

\author{%
  Xuan Li \\
  Amazon \\
  \texttt{milanlx@amazon.com} \\
  \And
  Yun Wang \\ 
  Amazon \\
  \texttt{yunwng@amazon.com} \\
  \And
  Bo Li \\
  Amazon \\
  \texttt{booli@amazon.com} \\
}
  
\begin{document}

\maketitle

\begin{abstract}
Tabular neural network (NN) has attracted remarkable attentions and its recent advances have gradually narrowed the performance gap with respect to tree-based models on many public datasets. While the mainstreams focus on calibrating NN to fit tabular data, we emphasize the importance of homogeneous embeddings and alternately concentrate on regularizing tabular inputs through supervised pretraining. Specifically, we extend a recent work (DeepTLF \cite{borisov2022deeptlf}) and utilize the structure of pretrained tree ensembles to transform raw variables into a single vector (T2V), or an array of tokens (T2T). Without loss of space efficiency, these binarized embeddings can be consumed by canonical tabular NN with fully-connected or attention-based building blocks. Through quantitative experiments on 88 OpenML datasets with binary classification task, we validated that the proposed tree-regularized representation not only tapers the difference with respect to tree-based models, but also achieves on-par and better performance when compared with advanced NN models. Most importantly, it possesses better robustness and can be easily scaled and generalized as standalone encoder for tabular modality. Codes: \href{https://github.com/milanlx/tree-regularized-embedding}{https://github.com/milanlx/tree-regularized-embedding}. 

\end{abstract}

%% ------- introduction ------- %%
% --------------                                                     
\section{Introduction}

% --motivation
Neural Network has achieved exceptional breakthroughs in the unstructured data regimes including image \cite{dosovitskiy2020image, oquab2023dinov2}, text \cite{brown2020language, radford2021learning}, video \cite{liu2022video, singer2022make} and speech \cite{baevski2020wav2vec, yang2023uniaudio}, whereas its performance is still capped by tree-based approaches when applied to structured tabular datasets \cite{grinsztajn2022tree, mcelfresh2023neural}. As there are growing demands on leveraging NN's capability to incorporate tabular modality for broader use cases such as multimodal learning \cite{ebrahimi2023lanistr, erickson2022multimodal, hager2023best, shi2021benchmarking, zhang2023meta}, it is critical to further boost tabular NN to its upper limit to better support these expansions. 

% --current practice and limitation 
Many recent works have attempted to bridge this gap by applying techniques that have demonstrated superior performance on other modalities to tabular learning. For example, a majority of the approaches follow a model-centric paradigm of applying simple feature transformation yet sophisticated customization on NN frameworks to fit tabular input. However, the underemphasis on feature quality could overshadow the efficacy of NN. Essentially, unlike image, text and speech data which have basic units (pixel, word, phoneme) that formulate a homogeneous representation space, tabular features are heterogeneous in nature as the columns possess different data sources, scales and distributions \cite{agarwal2021neural, fiedler2021simple, ma2020vaem}. Likewise, simple feature transformations such as min-max normalization might be incapable to make tabular input homogeneous enough to be consumed by NN backbones. Subsequently, we follow the data-centric scenario and seek data transformation strategies to acquire dedicated tabular embeddings.  

\begin{figure}[h!]
\centering
\includegraphics[width=.7\linewidth]{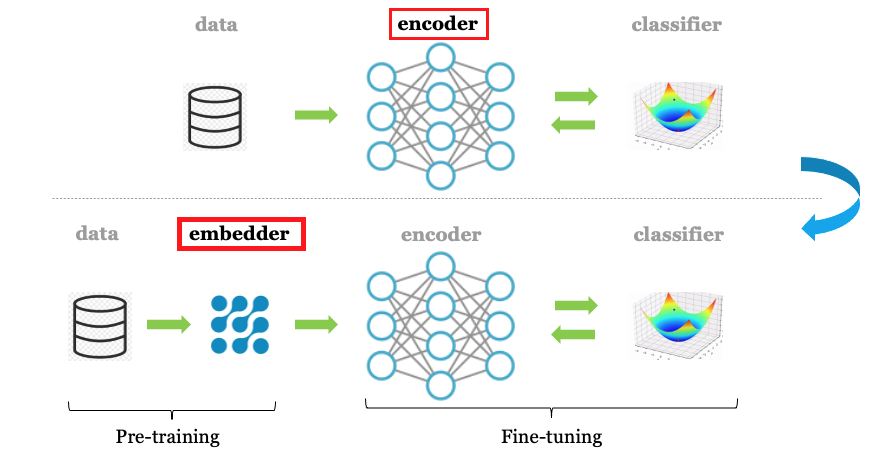}
\caption{An overview of data-centric tabular learning}
\label{fig:paradigm_shift}
\end{figure}

% --introduce method and novelty 
%-- high-level description of the contributions and results
Precisely, in this work we revisit the underexplored rationale on calibrating tabular data to fit NN. As visioned in Figure \ref{fig:paradigm_shift}, we leverage supervised pretraining to learn tree-regularized representations through an embedder module. In a snapshot, the proposed methodology exploits the structure of pretrained tree ensembles to generate binarized embeddings through a pairwise comparison between value in raw variable and the corresponding thresholds in tree node. Spanning the latent space of trees, the enriched representations can be fed into tabular NN directly and finetuned for different downstream tasks. In terms of implementation, we optimized and extended DeepTLF \cite{borisov2022deeptlf}, an overlooked advancement in boosting tabular NN with tree-transformed vector, to make it scalable for larger datasets and generalizable for vaster frameworks. On one hand, instead of transforming the data and storing the vectors all at once, we deploy it on-the-fly for each mini-batch during model training and inference, thus requesting no exhaustive memory usage. To compensate for the ensuing time complexity, we reformulate the pairwise comparison with matrix manipulation, which maintains the forward evaluation time at a similar scale. These two optimizations are essential for industrial tabular applications where the datasets might contain hundreds of columns and millions of rows. On the other hand, beyond generating embeddings as a single vector, we also treat each tree as tokenizer and further support tree-level transformation to obtain embeddings as an array of tokens. Essentially, it enables the representations to be compatible with attention-based  models \cite{huang2020tabtransformer, somepalli2021saint} that have received increasing attentions in the tabular learning communities. For evaluation, we leverage the TabZilla framework \cite{mcelfresh2023neural} and compare with a variety of state-of-the-art (SOTA) methods on 88 OpenML datasets with binary classification tasks. 

In summary, the contributions and novelties of this work are as follows:

\begin{itemize}
    \item We approach tabular representation learning from a data-centric perspective. Through a toy synthetic experiment, we reveal that simple NN model can always outperform well-tuned tree-based model in a homogeneous space, and therefore highlight the desideratum of tabular-specific transformations. 
    
    \item We improve a recent approach,  DeepTLF \cite{borisov2022deeptlf}, and further implement scalable algorithms to obtain tree-regularized tabular embeddings as a single vector (T2V), or an array of vectors (T2T). In essence, the transformed representations can be directly integrated with advanced tabular NN models with multi-layered perception (MLP) or multi-head attention (MHA) as building blocks.
    
    \item 
    We run comprehensive evaluations with a collection of 88 OpenML datasets on binary classification tasks. We validate that T2T with MHA backbones can narrow the performance gap with respect to tree-based models and achieve comparable or better performance compared to SOTA tabular NN models. More importantly,
    our methods show better robustness, and support generalizations at scale.
\end{itemize}

%% ------- related work ------- %%
% --------------    
\section{Related Work}

% ---- Heterogeneity
\paragraph{Heterogeneity in tabular emebddings} Unlike image, text and speech data that are composed of homogeneous units such as pixel, character and spectral band, tabular data are usually gathered from various information sources which made it heterogeneous by design. For example, tabular variables have different distributions \cite{ma2020vaem}, locate in irregular spaces \cite{ma2020vaem}, and contain different types including categorical, numerical and ordinal \cite{agarwal2021neural, tancik2020fourier} format. Although several researchers \cite{fiedler2021simple, ma2020vaem} have pointed out heterogeneity to be the fundamental blocker that restricts NN's generalization on tabular data, qualitative definitions and quantitative metrics are still missing for rigorous evaluations. However, the t-SNE plots \cite{van2008visualizing} can be utilized as a qualitative proxy to visualize the level of heterogeneity for different tabular representations \cite{borisov2022deeptlf}.  

% ---- Tabular NN
\paragraph{Tabular NN models and pretraining} Inspired by the recent advance of NN in other fields, many researchers have customized these techniques for tabular modality from two perspectives including modeling architectures and pretraining frameworks. 

In terms of modeling architectures, MLP \cite{fiedler2021simple, gorishniy2021revisiting, gorishniy2022embeddings, kadra2021well}, MHA \cite{chen2023trompt, gorishniy2021revisiting, gorishniy2022embeddings, huang2020tabtransformer, somepalli2021saint}, CNN \cite{zhu2021converting} and GNN \cite{du2021tabularnet} have been modified and found effective to boost performance over tree models on different public datasets. Although there is still no single option that dominates the rest, there are growing interests of adapting MHA in recent progress such as multimodal learning \cite{ebrahimi2023lanistr} and reasoning with language models \cite{hegselmann2023tabllm}. Intuitively, the self-attention mechanism in MHA is designed to discover relational pattern among the input features, i.e., understanding the context between words, which is similar to the conditional split mechanism utilized in tree-based models.

Besides, unsupervised, self-supervised and supervised pretraining have been leveraged by many works to obtain tabular-specific embeddings. For unsupervised scenario, quantile binning and periodic activation have been explored to independently encode each feature without interactions \cite{gorishniy2022embeddings}. For self-supervised pretext tasks, contrastive learning \cite{bahri2021scarf, chen2023recontab, darabi2021contrastive, hager2023best, rubachev2022revisiting, somepalli2021saint, yoon2020vime} and masked reconstruction \cite{arik2021tabnet, huang2020tabtransformer, majmundar2022met, rubachev2022revisiting, ucar2021subtab, yoon2020vime} are commonly adopted and the latter is reported to have better performance. For the supervised counterpart, knowledge distillation from ensembles of pretrained NNs \cite{lee2023practical} or boosting trees \cite{borisov2022deeptlf, ke2019deepgbm, wang2018tem} are implemented and reported to outperform tree models. However, this array of research is not well-explored, which is probably due to the concerns of overfitting \cite{feng2021rethinking} and scalability.

% --------------    
\section{Towards Data-Centric Tabular Learning}
In contrast to model-centric approaches that focus on calibrating NN models to fit with tabular data, we highlight the coupling effect between homogeneous features and NN models, and instead leverage pretraining to regularize the input latent space. As showed in Figure \ref{fig:paradigm_shift}, we first utilize an embedder at pretraining stage to learn representations through supervised pretraining. Specifically, we implement tree-to-vector (T2V) to support fully-connected encoders, and tree-to-tokens (T2T) to support attention-based encoders. Before diving into the technical details, we first introduce a synthetic experiment that motivates us towards doubling down on data-centric approaches. 

\paragraph{notations}
Let $\R^n$ be the $n$-dimensional Euclidean space and $||\cdot||_2$ be the Euclidean norm (L2 norm). We denote the unit hypersphere in $\R^d$ by $\mathbb{S}^{d-1} := \{ \mathrm{x} \in \R^d \: : \:  ||\mathrm{x}||_2 = 1\}$. We use $f_{\theta}(\cdot)$ to denote function $\{f(\cdot)\: : \: \R^d \rightarrow \R^c \}$ parameterized by $\theta$. With loss of generality, we use $x, \mathrm{x}, X$ to represent scalar, vector and matrix respectively. For matrix $X$, we use $X_i^j$ to index the element in the $i$-th row and $j$-th column.

% --------
\subsection{Synthetic Experiments}
\begin{figure}[h]
\begin{floatrow}
\ffigbox{%
  \includegraphics[scale=0.30]{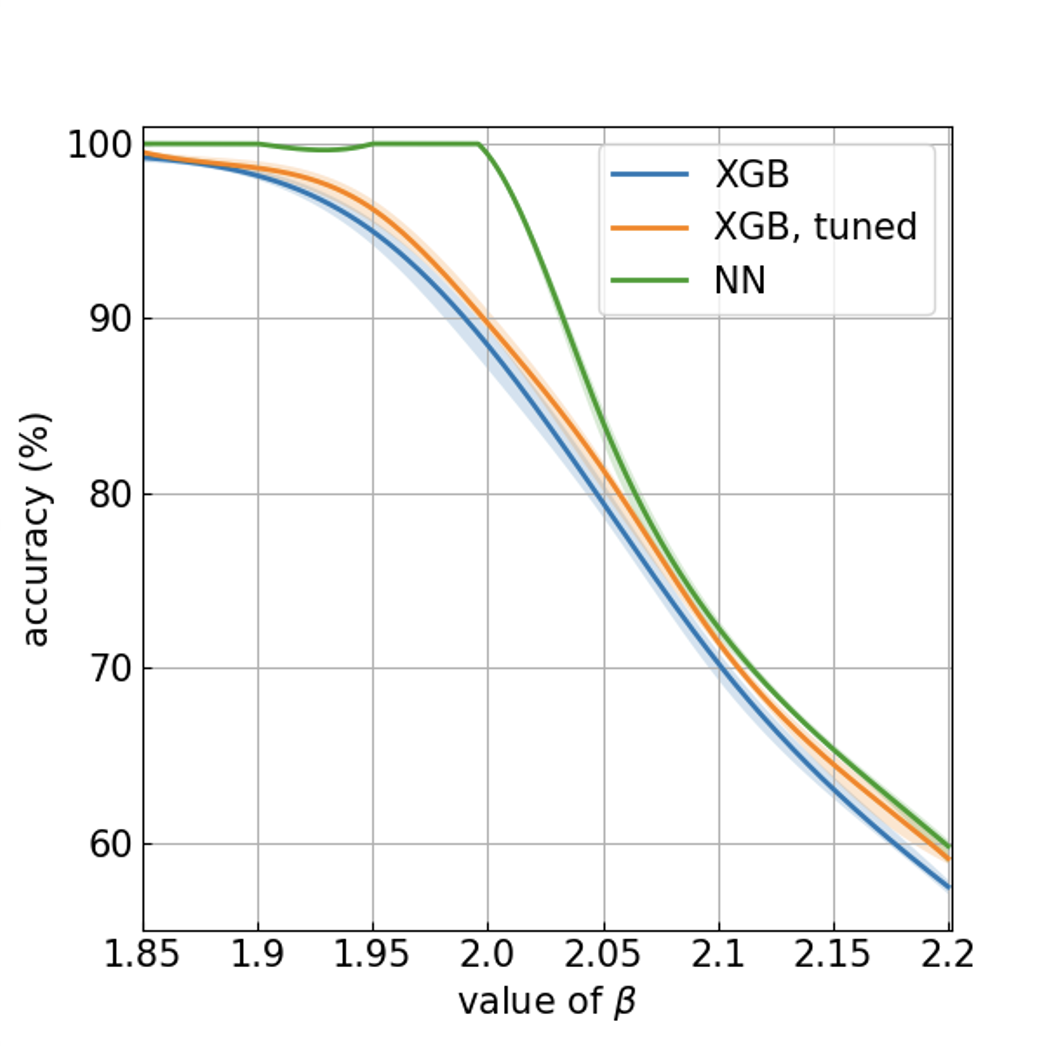}
}{%
  \caption{Comparison between MLP and XGB with varying $\beta$ in terms of accuracy}%
  \label{fig:synthetic_result}
}
\ffigbox{%
  \includegraphics[scale=0.30]{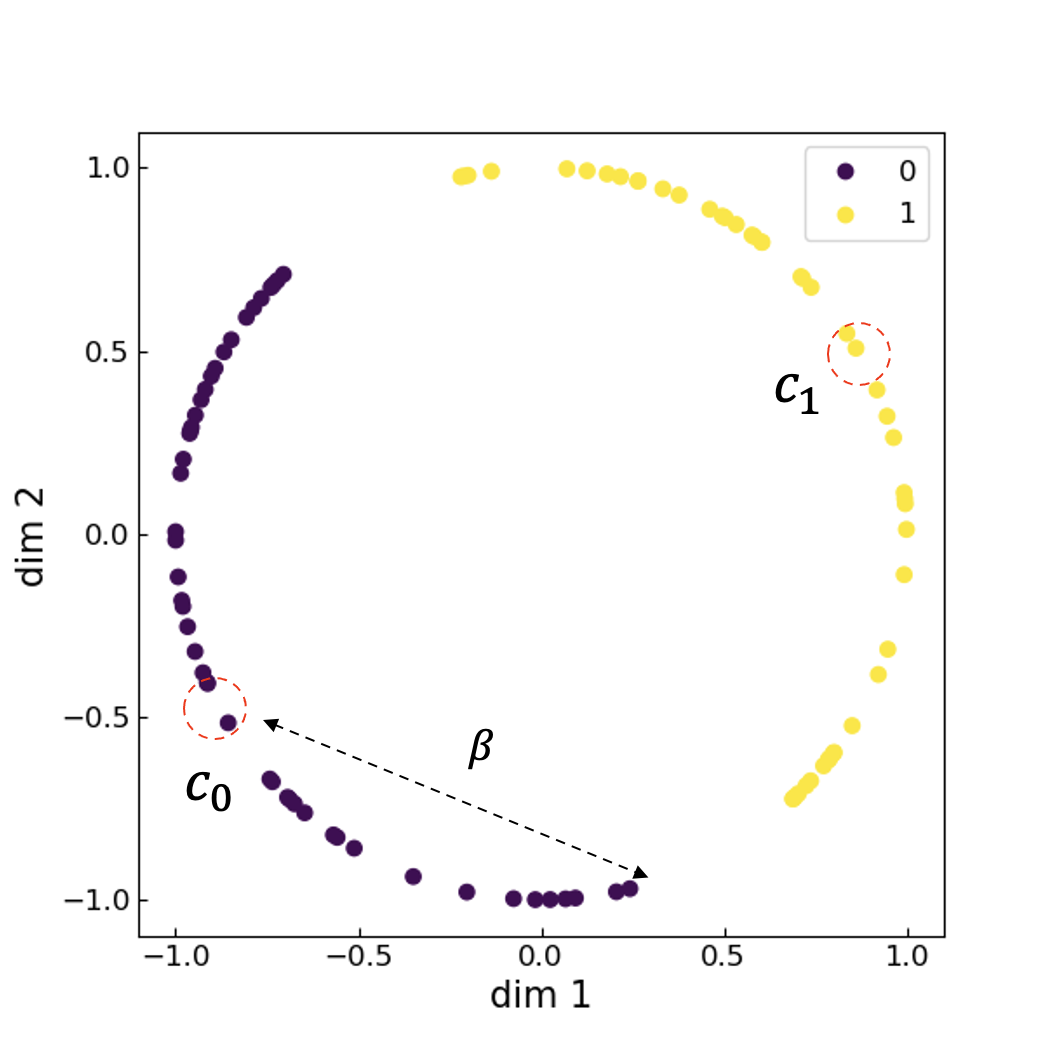}
}{%
  \caption{A visualization of the synthetic data in 2D scenario}%
  \label{fig:synthetic_2d}
}
\end{floatrow}
\end{figure}

To validate the coupling effects between homogeneous latent space and neural models, we conduct a toy experiment with synthetic data which simulates homogeneous feature spaces. For this homogeneous scenario, we generate balanced 100-dimensional data that are uniformly pinpointed on a unit hypersphere around two central points $\mathrm{c_0}$ and $\mathrm{c_1}$, where the two centers are diagonal to each other and also are located on that unit hypersphere, i.e., $\mathrm{c_0} = - \mathrm{c_1}$.  We use the term $\beta$ to control the maximum distance between a sample $(\mathrm{x}, y)$ and its central point, i.e., $P(y = i \: | \: ||\mathrm{x} - \mathrm{c_i}||_2 \leq \beta) = 1$. Intuitively, a small $\beta$ indicates the data are tightly clustered around centers, while a large $\beta$ indicates patterned overlapping on the boundaries. An illustrative visualization of the synthetic data in 2-dimensional scenario can be found in Figure \ref{fig:synthetic_2d}. 

Through uniform sampling with rejection, we generate 10k balanced samples and split them into training, validation and testing bucket with 60\%, 20\% and 20\% in proportion. For comparison, we train a two-layer MLP ($100 \rightarrow 100  \rightarrow 2$) as NN model, a XGBoost (XGB) with default hyperparameter, and a XGB with well-tuned hyperparameter as tree-based models. We run 5 trials of experiment per $\beta$ and report the average of accuracy in Figure \ref{fig:synthetic_result}. By varying $\beta$ between $1.85$ and $2.20$ with a $0.05$ interval, we found that NN can always outperform the default as well as the well-tuned XGB in this hyperspherical feature space. With different features regularized within the same scale, we posit NN might have superiority over tree-based models in this homogeneous latent space, and therefore introduce tree-regularized embeddings that are aligned with this observation.

% ------
\subsection{Tree-regularized Embedding}

\begin{figure}[h!]
\centering
\includegraphics[width=.4\linewidth]{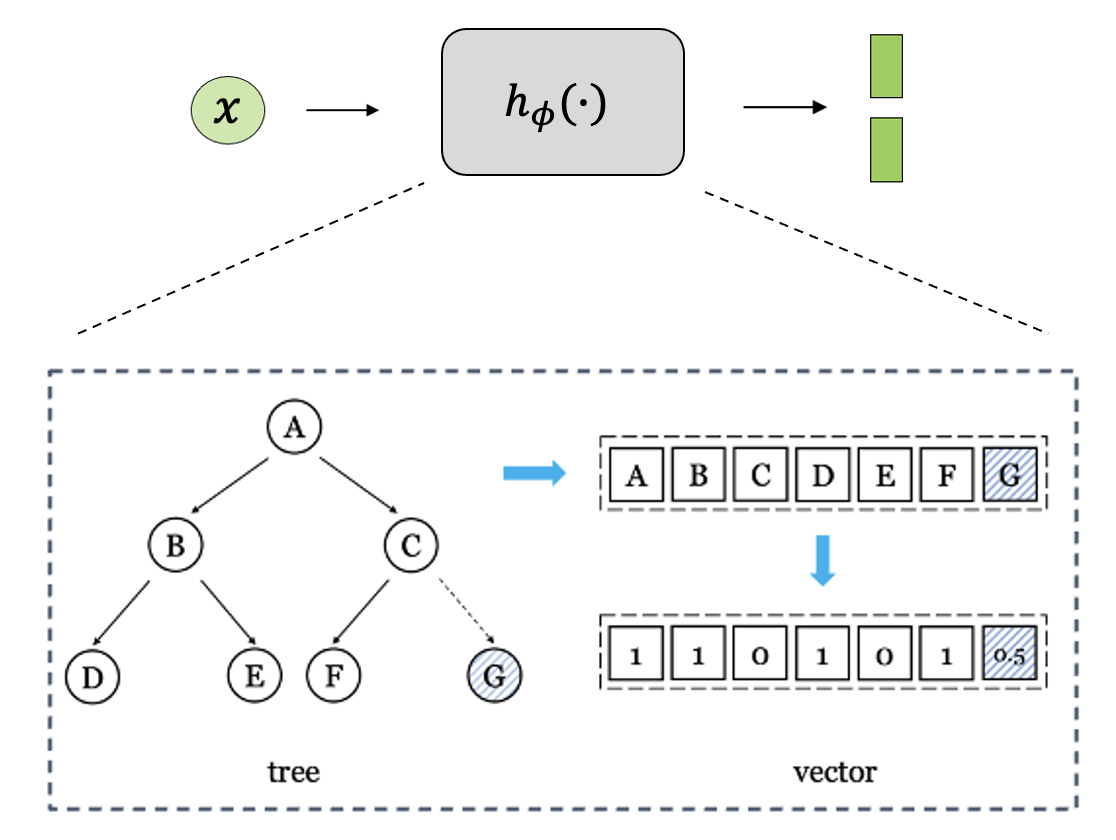}
\caption{Overview of tree-to-vector (T2V) embedding }
\label{fig:overview_t2v}
\end{figure}

\paragraph{supervised tree-regularized embeddings}
As a realization of supervised pretraining, the tree-regularized approach takes advantages of tree information from XGB to formulate new embeddings with feature interactions. Ideally, this procedure will transform the heterogeneous tabular data into homogeneous format by distilling knowledge from nodes of trained decision trees \cite{borisov2022deeptlf}. As showed in Figure \ref{fig:overview_t2v}, it will firstly extracts node information - a tuple of variable index and threshold - from each tree as a map, and then binarizes each data by comparing the corresponding variable value with respect to the threshold given the index. Interested readers can refer to Figure \ref{fig:t2v_example} for an illustrative example. To make the embedder compatible with different NN encoders and scalable with large datasets, we extend this simple setup from work \cite{borisov2022deeptlf} and introduce T2V and T2T to support fully-connected and attention-based models. 

\begin{figure}[h]
\centering
\includegraphics[width=1.0\linewidth]{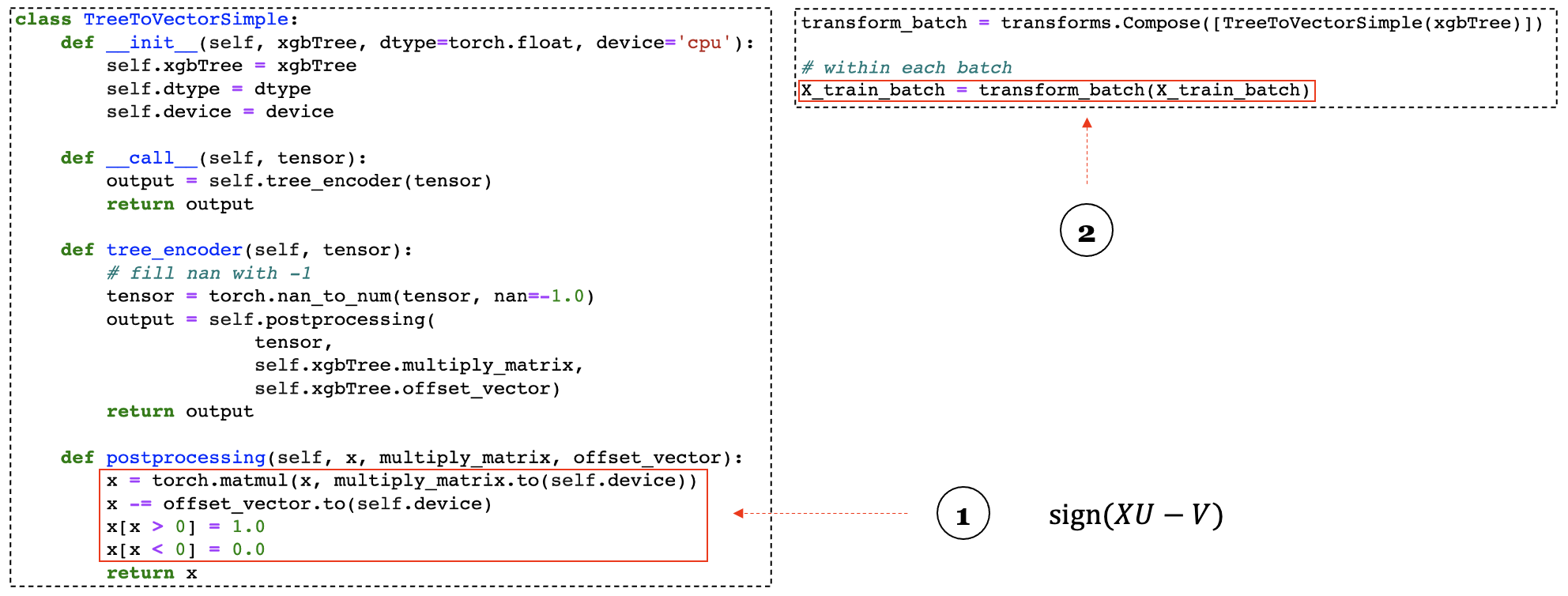}
\caption{Pseudocode of in-batch transformation for T2V in a PyTorch-like style. Step 1 replaces pairwise comparison with matrix manipulation, while Step 2 showcases on-the-fly transformations for mini-batch implemented through the $\mathrm{tranforms.Compose}$ module in PyTorch.} 
\label{fig:in_batch_transformation}
\end{figure}

\textbf{T2V}: With the embedding vectors extracted from each tree, we perform a preprocessing on the collection of  $\{\mathrm{variable\_ index: threshold}\}$ map to remove duplicated instances based on rounded threshold, concatenate the vectors to form a single one-dimensional vector, and finally integrate the embedding with MLP encoders during model training. To make the embedder scalable, we reformulate the pairwise (\{value, threshold\}) comparison with matrix manipulation, and only employ this operation within each mini-batch on the fly, which we denote as in-batch transformation. Specifically, assume we have a data matrix $X \in \R^{n\times m}$ with $n$ instances and $m$ variables, and a corresponding collection $M \in \R^{k\times 2}$ with $k$ pairs of the $\{\mathrm{variable\_ index, threshold}\}$ map extracted from tree ensembles (XGB). According to Eq (\ref{eq:t2v}), we can construct a matrix $U \in \R^{m\times k}$, and a matrix $V \in \R^{m\times k}$ composed of $m$ stacked vector $\mathrm{v}$ ($ \mathrm{v}\in \R^{k}, \mathrm{v}_i = M_i^2$), so that the operation of  $\mathrm{sign}(X U - V)$ is equivalent to the iterative pairwise comparison of \{value, threshold\}. Most importantly, the in-batch transformation makes the algorithm generalizable to much larger datasets with hundreds of columns and millions of rows. We provide the details in Algorithm (\ref{alg:t2v_I}) and a PyTorch-like pseudocode in Figure \ref{fig:in_batch_transformation}. 

\begin{equation} \label{eq:t2v}
    U_{M_{i}^1}^{i} = 
    \begin{cases}
        \:\: 1, & \forall \: i \in \{1, 2, ..., k\} \\
        \:\: 0, & \text{otherwise}
    \end{cases} 
\end{equation}

\textbf{T2T}: To make it compatible with MHA backbone, we treat the embeddings from each tree as token and apply paddings to ensure every token are aligned in dimension. The final embeddings for each data instance have a dimension of $\R^{d \times k}$, where $d$ is the number of tree ensembles in XGB and $k$ is the maximum number of nodes in these trees. Precisely, we pad $0.5$ to non-splitting nodes (to make tree complete) and $-1.0$ at the tail of the embedding vector to make it aligned with dimension $k$. To ensure the semantics of token are consistent, we preserve the topological order of each tree through level order traversal when extracting tree nodes. The details of these operations can be found in Algorithm (\ref{alg:t2v_II}) and Figure \ref{fig:t2v_flowchart}. Matrix manipulations and in-batch transformation are applied similarly as T2V to account for scalability. Intuitively, the final output $X$ ($X \in \R^{n\times d \times k}$) can be regarded as an array of tokens and directly consumed by transformers with attention block.

% --------------    
\section{Experiments}

% -------- training details 
\subsection{Datasets, models, and training details }
We leverage a subset of the benckmark datasets provided in TabZilla \cite{mcelfresh2023neural} repository to evaluate the effectiveness, generalizability and scalability of the proposed methods. Specifically, we select 91 OpenML \footnote{https://www.openml.org/} datasets with binary classification task and utilize the Area Under the Curve (AUC) in percentage as evaluation metrics. We apply light preprocessing to fill missing value with zero and convert categorical variables to ordinal values through label encoding. 

We keep model framework consistent throughout the experiments. For T2V, we use two-layered MLP with ReLU activation and fix the hidden dimensions as $m \rightarrow 256 \rightarrow 128  \rightarrow 2$, where $m$ is the dimension of T2V embeddings. For T2T, we use MHA encoder configured with 2 identical building blocks, where each block consists of 4 heads with embedding dimension as 8. An one-layered MLP ($m \rightarrow 128  \rightarrow 2$) is connected with the concatenated output of MHA as classification head. For comprehensive comparisons, we select CatBoost \cite{prokhorenkova2018catboost}, XGBoost \cite{chen2015xgboost} and LightGBM \cite{ke2017lightgbm} as tree-based baselines. In addition, we use SAINT \cite{somepalli2021saint} and the ResNet-like model \cite{gorishniy2021revisiting} as SOTA NN baselines given the rankings reported in \cite{mcelfresh2023neural}. Finally, we include a two-layered MLP ($m \rightarrow 128  \rightarrow 2$, denoted as MLP) with min-max normalization applied on raw variables as a vanilla NN baseline.  

For evaluation, we leverage the default 10 training/testing splits provided by OpenML and report the mean AUC over the 10 runs for each dataset. Similar to TabZilla, for each split we further extract a fixed validation set from the training set to make the training/validation/testing proportion as 80\%, 10\% and 10\% respectively. Additionally, we fix the hyperparameters for each model with their default values for generalization purpose. Specifically, for all NN-based models we apply Adam as default optimizer with learning rate as $0.001$ and batch size as 64. Early stopping with 10 epochs and 600 seconds timeout is applied to both tree-based and NN-based models. All experiments are run on an A10G GPU with approximately 3 GPU days.

% -------- performance details 
\subsection{Performance Evaluation}
We summarize the experiment results in this section. In terms of robustness, we find most of the NN models cannot generalize to the entire datasets, and therefore compare models in full-scale and partial-scale scenarios based on their dataset coverage. Precisely, we compare T2V, T2T, MLP with tree-based models on 88 datasets as full-scale scenario. For partial-scale case, we compare T2V with SAINT and ResNet on 59 and 73 datasets respectively. Also, we provide a heuristic analysis on the time complexity of in-batch transformation by varying batch size and number of tree ensembles. 

% ----
\paragraph{robustness}
We report the number of datasets that can be evaluated by each method in Table \ref{tab:dataset_coverage}. In general, we find tree-based models achieve the best robustness while NN models, such as SAINT and ResNet, suffer from numerical and timeout issue on a variety of datasets. Notably, T2V and T2T have better robustness as they can generalize to 88/91 of the cases. 

% data coverage 
\begin{table}[h]
    \centering
    \begin{tabular}{c|c|c|c|c|c|c|c}
    \toprule 
    CatBoost & XGBoost & LightGBM & T2V & T2T & MLP & SAINT & ResNet \\
    \midrule 
    91 & 91 & 91 & 88 & 88 & 88 & 59 & 73 \\
    \bottomrule
    \end{tabular}
    \caption{Number of datasets can be evaluated by tree-based and NN-based models}
    \label{tab:dataset_coverage}
\end{table}

% ----
\paragraph{full-scale comparison}
Given the availability of data coverage, we first compare T2V, T2T and the vanilla MLP with respect to tree-based models. The results are reported in Table \ref{tab:rank_result} where the methods are ranked by the mean AUC taken over across the 88 overlapped datasets. The distribution of AUC attained by different method is showed in Figure \ref{fig:boxplot_full_scale} in Appendix. Firstly, while T2T outperforms the vanilla MLP, it still has a $3.43\%$ gap in percentaged AUC with respect to the best tree-based model. Second, T2V underperforms MLP, probably because a shallow NN backbone is not sufficient for the high-dimensional embeddings. Moreover, we point out the diversity existed in the datasets as each method can achieve the highest as well as the lowest ranking. This observation is aligned with the results reported in TabZilla \cite{mcelfresh2023neural}, where the authors found no single approach can consistently dominate the rest and the difference in performance was insignificant in many of the cases. 

\begin{table}[h]
    \centering
    \begin{tabular}{c|c c c c|c}
    \toprule 
    Algorithm & \multicolumn{4}{|c|}{Rank $\downarrow$} & AUC (\%) $\uparrow$ \\ 
    \midrule
     & min & max & mean & median & mean \\
    \midrule 
    CatBoost & $1$ & $6$ & $2.38$ & $2$ & $88.06$ \\
    \midrule 
    XGBoost & $1$ & $6$ & $2.83$ & $2$ & $87.70$ \\
    \midrule
    LightGBM & $1$ & $6$ & $3.16$ & $3$ & $86.37$ \\
    \midrule
    T2T & $1$ & $6$ & $4.07$ & $4$ & $84.63$ \\
    \midrule
    MLP & $1$ & $6$ & $4.22$ & $4$ & $84.42$ \\
    \midrule
    T2V & $1$ & $6$ & $4.45$ & $5$ & $83.15$ \\
    \bottomrule
    \end{tabular}
    \caption{Comparison between T2V, T2T, MLP and tree-based models on 88 datasets}
    \label{tab:rank_result}
\end{table}

\begin{figure}[h]
\begin{floatrow}
\ffigbox{%
  \includegraphics[scale=0.4]{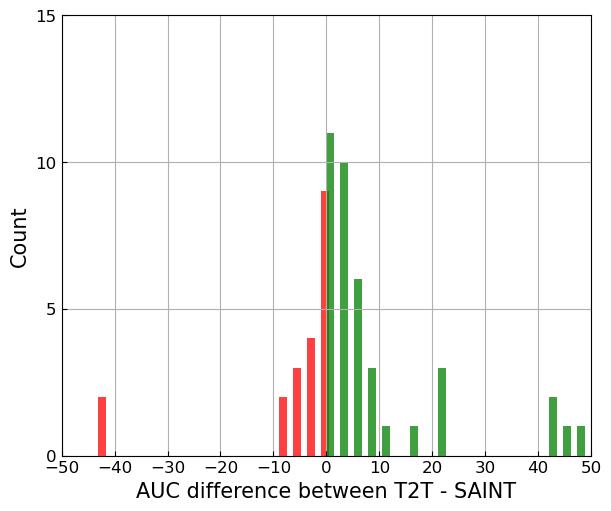}
}{%
  \caption{Histogram of difference in AUC between T2T and SAINT}%
  \label{fig:compare_t2t_and_saint}
} 
\ffigbox{%
  \includegraphics[scale=0.4]{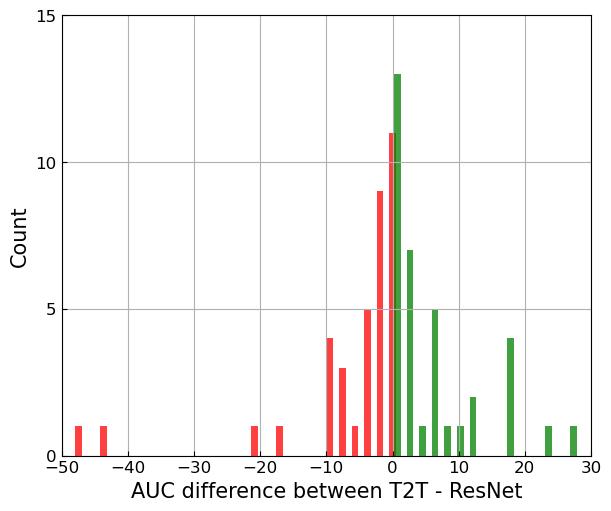}
}{%
  \caption{Histogram of difference in AUC between T2T and ResNet}%
  \label{fig:compare_t2t_and_resnet}
}
\end{floatrow}
\end{figure}

% ----
\paragraph{partial-scale comparison} 
Given the results from full-scale comparison, we also conduct pairwise comparison between T2T, SAINT and ResNet on the intersected datasets. For comparison, we check the difference in percentaged AUC between two methods and define a win on a dataset if the former method achieves a high AUC. The histogram of difference in AUC between \{T2T, SAINT\} and \{T2T, ResNet\} are showed in Figure \ref{fig:compare_t2t_and_saint} and \ref{fig:compare_t2t_and_resnet} respectively. Comparing T2T and SAINT, we find the former win 39 out of 59 of the datasets ($66.10\%$) and achieve a $3.74\%$ absolute lift in percentaged AUC. When compared with ResNet, however, we find T2T can win 36 of the 73 cases ($49.31\%$) with a $0.13\%$ difference in percentaged AUC on average. From the histogram it is found the majority of the differences are within $0\% - 10\%$ range, and each method has generalization issue on several datasets. The distribution of the AUC can be found in Figure \ref{fig:boxplot_partial_scale}.

\begin{figure}[h]
\centering
\includegraphics[width=.5\linewidth]{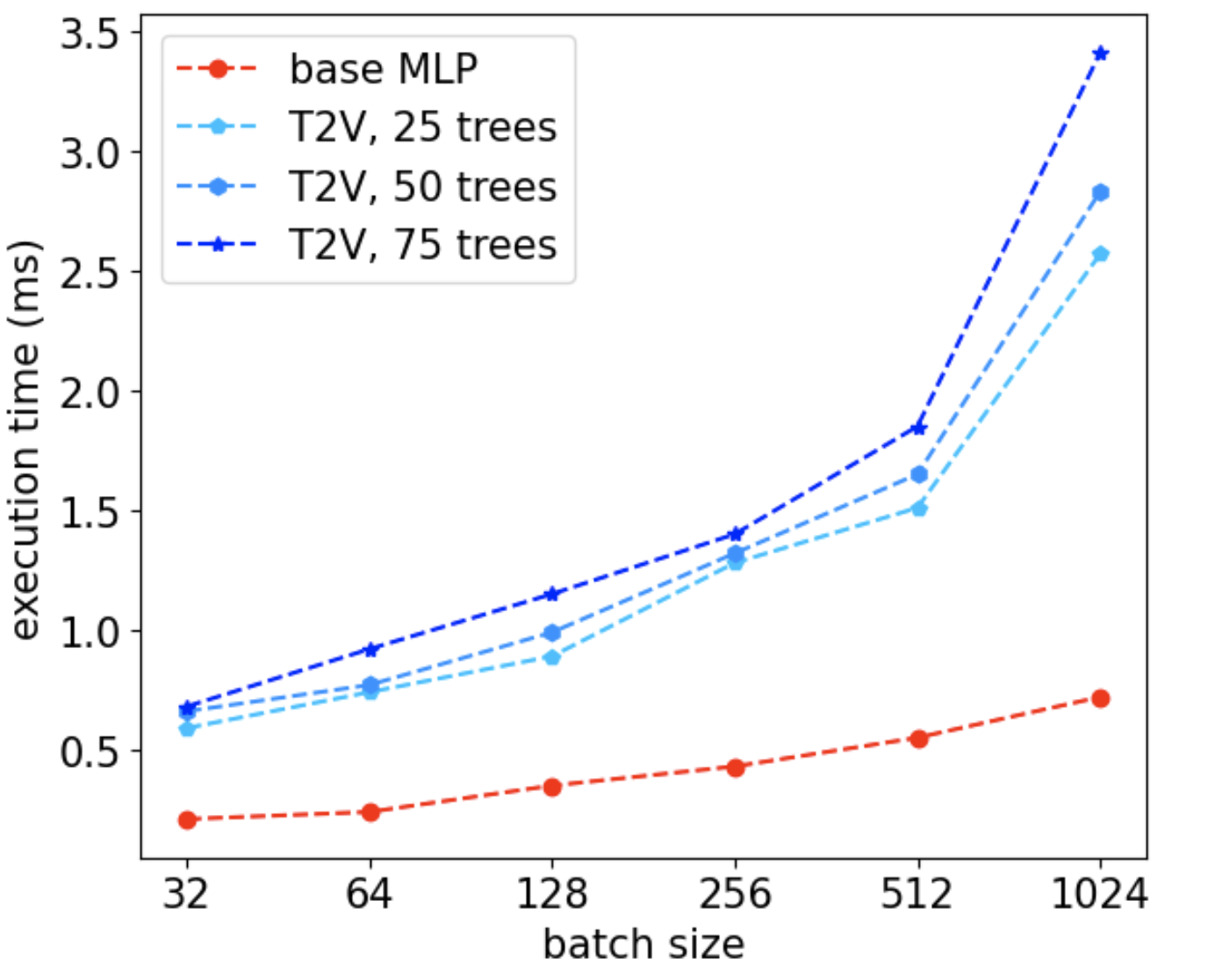}
\caption{Comparison of time complexity between T2V and vanilla MLP on synthetic datasets}
\label{fig:t2v_time_complexity}
\end{figure}

% ----
\paragraph{time complexity analysis} 
As our methods made a trade-off between time and space complexity, we further conduct an analysis to evaluate the computational overhead with the synthetic datasets introduced in the previous section. Basically, we compare the forward-pass time between T2V with MLP and vanilla MLP for mini-batch evaluations. The results are showed in Figure \ref{fig:t2v_time_complexity}, where the execution time is reported as the average over 10 runs per scenario. By varying the batch size and number of tree ensembles, we find T2V scales well with respect to number of tree ensembles. However, for each mini-batch it takes 3x - 5x evaluation time when compared to the vanilla MLP for batch size up to 512.

% --------------    
\section{Conclusions and Future Works}
We follow a data-centric perspective and propose two methods to obtain tree-regularized embeddings with efficient in-batch transformation. Our improved tabular embeddings, T2V and T2T, can be simply consumed by many tabular NN frameworks with MLP and MHA as building block. Through comprehensive evaluations on 88 OpenML datasets, we show strong robustness and on-par performance with respect to SOTA NN models on binary classification tasks. These results demonstrate the potential of generalizing and scaling our approaches as tabular encoder for broader applications that require tabular modality. 

We plan to explore several directions to further improve the effectiveness and scalability of the proposed methods. Firstly, we will conduct architecture search to explore consonant NN designs that works with tree-regularized embeddings. In addition, for T2T we will try to further encode each tree as discrete token and utilize self-supervised pretraining to learn embeddings with customizable dimension through contrastive or reconstruction task. Finally, we point out a lack of quantitative metric on homogeneity and benchmark datasets at industrial scale, which are worth exploring in the next sprint.

% --------------    
\section*{Acknowledgements}
We would like to thank Ege Beyazit, Jonathan Kozaczuk, Mihir Pendse, Pankaj Rajak, Jiajian Lu and Vanessa Wallace for valuable discussions, feedback and support.

%\newpage
\bibliography{ref.bib} 

\begin{thebibliography}{10}

\bibitem{agarwal2021neural}
Rishabh Agarwal, Levi Melnick, Nicholas Frosst, Xuezhou Zhang, Ben Lengerich,
  Rich Caruana, and Geoffrey~E Hinton.
\newblock Neural additive models: Interpretable machine learning with neural
  nets.
\newblock {\em Advances in Neural Information Processing Systems},
  34:4699--4711, 2021.

\bibitem{arik2021tabnet}
Sercan~{\"O} Arik and Tomas Pfister.
\newblock Tabnet: Attentive interpretable tabular learning.
\newblock In {\em Proceedings of the AAAI Conference on Artificial
  Intelligence}, volume~35, pages 6679--6687, 2021.

\bibitem{baevski2020wav2vec}
Alexei Baevski, Yuhao Zhou, Abdelrahman Mohamed, and Michael Auli.
\newblock wav2vec 2.0: A framework for self-supervised learning of speech
  representations.
\newblock {\em Advances in Neural Information Processing Systems},
  33:12449--12460, 2020.

\bibitem{bahri2021scarf}
Dara Bahri, Heinrich Jiang, Yi~Tay, and Donald Metzler.
\newblock Scarf: Self-supervised contrastive learning using random feature
  corruption.
\newblock {\em arXiv preprint arXiv:2106.15147}, 2021.

\bibitem{borisov2022deeptlf}
Vadim Borisov, Klaus Broelemann, Enkelejda Kasneci, and Gjergji Kasneci.
\newblock Deeptlf: robust deep neural networks for heterogeneous tabular data.
\newblock {\em International Journal of Data Science and Analytics}, pages
  1--16, 2022.

\bibitem{brown2020language}
Tom Brown, Benjamin Mann, Nick Ryder, Melanie Subbiah, Jared~D Kaplan, Prafulla
  Dhariwal, Arvind Neelakantan, Pranav Shyam, Girish Sastry, Amanda Askell,
  et~al.
\newblock Language models are few-shot learners.
\newblock {\em Advances in neural information processing systems},
  33:1877--1901, 2020.

\bibitem{chen2023trompt}
Kuan-Yu Chen, Ping-Han Chiang, Hsin-Rung Chou, Ting-Wei Chen, and Tien-Hao
  Chang.
\newblock Trompt: Towards a better deep neural network for tabular data.
\newblock {\em arXiv preprint arXiv:2305.18446}, 2023.

\bibitem{chen2023recontab}
Suiyao Chen, Jing Wu, Naira Hovakimyan, and Handong Yao.
\newblock Recontab: Regularized contrastive representation learning for tabular
  data.
\newblock {\em arXiv preprint arXiv:2310.18541}, 2023.

\bibitem{chen2015xgboost}
Tianqi Chen, Tong He, Michael Benesty, Vadim Khotilovich, Yuan Tang, Hyunsu
  Cho, Kailong Chen, Rory Mitchell, Ignacio Cano, Tianyi Zhou, et~al.
\newblock Xgboost: extreme gradient boosting.
\newblock {\em R package version 0.4-2}, 1(4):1--4, 2015.

\bibitem{darabi2021contrastive}
Sajad Darabi, Shayan Fazeli, Ali Pazoki, Sriram Sankararaman, and Majid
  Sarrafzadeh.
\newblock Contrastive mixup: Self-and semi-supervised learning for tabular
  domain.
\newblock {\em arXiv preprint arXiv:2108.12296}, 2021.

\bibitem{dosovitskiy2020image}
Alexey Dosovitskiy, Lucas Beyer, Alexander Kolesnikov, Dirk Weissenborn,
  Xiaohua Zhai, Thomas Unterthiner, Mostafa Dehghani, Matthias Minderer, Georg
  Heigold, Sylvain Gelly, et~al.
\newblock An image is worth 16x16 words: Transformers for image recognition at
  scale.
\newblock {\em arXiv preprint arXiv:2010.11929}, 2020.

\bibitem{du2021tabularnet}
Lun Du, Fei Gao, Xu~Chen, Ran Jia, Junshan Wang, Jiang Zhang, Shi Han, and
  Dongmei Zhang.
\newblock Tabularnet: A neural network architecture for understanding semantic
  structures of tabular data.
\newblock In {\em Proceedings of the 27th ACM SIGKDD Conference on Knowledge
  Discovery \& Data Mining}, pages 322--331, 2021.

\bibitem{ebrahimi2023lanistr}
Sayna Ebrahimi, Sercan~O Arik, Yihe Dong, and Tomas Pfister.
\newblock Lanistr: Multimodal learning from structured and unstructured data.
\newblock {\em arXiv preprint arXiv:2305.16556}, 2023.

\bibitem{erickson2022multimodal}
Nick Erickson, Xingjian Shi, James Sharpnack, and Alexander Smola.
\newblock Multimodal automl for image, text and tabular data.
\newblock In {\em Proceedings of the 28th ACM SIGKDD Conference on Knowledge
  Discovery and Data Mining}, pages 4786--4787, 2022.

\bibitem{feng2021rethinking}
Yutong Feng, Jianwen Jiang, Mingqian Tang, Rong Jin, and Yue Gao.
\newblock Rethinking supervised pre-training for better downstream
  transferring.
\newblock {\em arXiv preprint arXiv:2110.06014}, 2021.

\bibitem{fiedler2021simple}
James Fiedler.
\newblock Simple modifications to improve tabular neural networks.
\newblock {\em arXiv preprint arXiv:2108.03214}, 2021.

\bibitem{gorishniy2022embeddings}
Yury Gorishniy, Ivan Rubachev, and Artem Babenko.
\newblock On embeddings for numerical features in tabular deep learning.
\newblock {\em Advances in Neural Information Processing Systems},
  35:24991--25004, 2022.

\bibitem{gorishniy2021revisiting}
Yury Gorishniy, Ivan Rubachev, Valentin Khrulkov, and Artem Babenko.
\newblock Revisiting deep learning models for tabular data.
\newblock {\em Advances in Neural Information Processing Systems}, 34, 2021.

\bibitem{grinsztajn2022tree}
L{\'e}o Grinsztajn, Edouard Oyallon, and Ga{\"e}l Varoquaux.
\newblock Why do tree-based models still outperform deep learning on tabular
  data?
\newblock {\em arXiv preprint arXiv:2207.08815}, 2022.

\bibitem{hager2023best}
Paul Hager, Martin~J Menten, and Daniel Rueckert.
\newblock Best of both worlds: Multimodal contrastive learning with tabular and
  imaging data.
\newblock In {\em Proceedings of the IEEE/CVF Conference on Computer Vision and
  Pattern Recognition}, pages 23924--23935, 2023.

\bibitem{hegselmann2023tabllm}
Stefan Hegselmann, Alejandro Buendia, Hunter Lang, Monica Agrawal, Xiaoyi
  Jiang, and David Sontag.
\newblock Tabllm: Few-shot classification of tabular data with large language
  models.
\newblock In {\em International Conference on Artificial Intelligence and
  Statistics}, pages 5549--5581. PMLR, 2023.

\bibitem{huang2020tabtransformer}
Xin Huang, Ashish Khetan, Milan Cvitkovic, and Zohar Karnin.
\newblock Tabtransformer: Tabular data modeling using contextual embeddings.
\newblock {\em arXiv preprint arXiv:2012.06678}, 2020.

\bibitem{kadra2021well}
Arlind Kadra, Marius Lindauer, Frank Hutter, and Josif Grabocka.
\newblock Well-tuned simple nets excel on tabular datasets.
\newblock {\em Advances in neural information processing systems},
  34:23928--23941, 2021.

\bibitem{ke2017lightgbm}
Guolin Ke, Qi~Meng, Thomas Finley, Taifeng Wang, Wei Chen, Weidong Ma, Qiwei
  Ye, and Tie-Yan Liu.
\newblock Lightgbm: A highly efficient gradient boosting decision tree.
\newblock {\em Advances in neural information processing systems}, 30, 2017.

\bibitem{ke2019deepgbm}
Guolin Ke, Zhenhui Xu, Jia Zhang, Jiang Bian, and Tie-Yan Liu.
\newblock Deepgbm: A deep learning framework distilled by gbdt for online
  prediction tasks.
\newblock In {\em Proceedings of the 25th ACM SIGKDD International Conference
  on Knowledge Discovery \& Data Mining}, pages 384--394, 2019.

\bibitem{lee2023practical}
Chung-Wei Lee, Pavlos~Anastasios Apostolopulos, and Igor~L Markov.
\newblock Practical knowledge distillation: Using dnns to beat dnns.
\newblock {\em arXiv preprint arXiv:2302.12360}, 2023.

\bibitem{liu2022video}
Ze~Liu, Jia Ning, Yue Cao, Yixuan Wei, Zheng Zhang, Stephen Lin, and Han Hu.
\newblock Video swin transformer.
\newblock In {\em Proceedings of the IEEE/CVF conference on computer vision and
  pattern recognition}, pages 3202--3211, 2022.

\bibitem{ma2020vaem}
Chao Ma, Sebastian Tschiatschek, Richard Turner, Jos{\'e}~Miguel
  Hern{\'a}ndez-Lobato, and Cheng Zhang.
\newblock Vaem: a deep generative model for heterogeneous mixed type data.
\newblock {\em Advances in Neural Information Processing Systems},
  33:11237--11247, 2020.

\bibitem{majmundar2022met}
Kushal Majmundar, Sachin Goyal, Praneeth Netrapalli, and Prateek Jain.
\newblock Met: Masked encoding for tabular data.
\newblock {\em arXiv preprint arXiv:2206.08564}, 2022.

\bibitem{mcelfresh2023neural}
Duncan McElfresh, Sujay Khandagale, Jonathan Valverde, Ganesh Ramakrishnan,
  Micah Goldblum, Colin White, et~al.
\newblock When do neural nets outperform boosted trees on tabular data?
\newblock {\em arXiv preprint arXiv:2305.02997}, 2023.

\bibitem{oquab2023dinov2}
Maxime Oquab, Timoth{\'e}e Darcet, Th{\'e}o Moutakanni, Huy Vo, Marc
  Szafraniec, Vasil Khalidov, Pierre Fernandez, Daniel Haziza, Francisco Massa,
  Alaaeldin El-Nouby, et~al.
\newblock Dinov2: Learning robust visual features without supervision.
\newblock {\em arXiv preprint arXiv:2304.07193}, 2023.

\bibitem{prokhorenkova2018catboost}
Liudmila Prokhorenkova, Gleb Gusev, Aleksandr Vorobev, Anna~Veronika Dorogush,
  and Andrey Gulin.
\newblock Catboost: unbiased boosting with categorical features.
\newblock {\em Advances in neural information processing systems}, 31, 2018.

\bibitem{radford2021learning}
Alec Radford, Jong~Wook Kim, Chris Hallacy, Aditya Ramesh, Gabriel Goh,
  Sandhini Agarwal, Girish Sastry, Amanda Askell, Pamela Mishkin, Jack Clark,
  et~al.
\newblock Learning transferable visual models from natural language
  supervision.
\newblock In {\em International Conference on Machine Learning}, pages
  8748--8763. PMLR, 2021.

\bibitem{rubachev2022revisiting}
Ivan Rubachev, Artem Alekberov, Yury Gorishniy, and Artem Babenko.
\newblock Revisiting pretraining objectives for tabular deep learning.
\newblock {\em arXiv preprint arXiv:2207.03208}, 2022.

\bibitem{shi2021benchmarking}
Xingjian Shi, Jonas Mueller, Nick Erickson, Mu~Li, and Alexander~J Smola.
\newblock Benchmarking multimodal automl for tabular data with text fields.
\newblock {\em arXiv preprint arXiv:2111.02705}, 2021.

\bibitem{singer2022make}
Uriel Singer, Adam Polyak, Thomas Hayes, Xi~Yin, Jie An, Songyang Zhang, Qiyuan
  Hu, Harry Yang, Oron Ashual, Oran Gafni, et~al.
\newblock Make-a-video: Text-to-video generation without text-video data.
\newblock {\em arXiv preprint arXiv:2209.14792}, 2022.

\bibitem{somepalli2021saint}
Gowthami Somepalli, Micah Goldblum, Avi Schwarzschild, C~Bayan Bruss, and Tom
  Goldstein.
\newblock Saint: Improved neural networks for tabular data via row attention
  and contrastive pre-training.
\newblock {\em arXiv preprint arXiv:2106.01342}, 2021.

\bibitem{tancik2020fourier}
Matthew Tancik, Pratul Srinivasan, Ben Mildenhall, Sara Fridovich-Keil, Nithin
  Raghavan, Utkarsh Singhal, Ravi Ramamoorthi, Jonathan Barron, and Ren Ng.
\newblock Fourier features let networks learn high frequency functions in low
  dimensional domains.
\newblock {\em Advances in Neural Information Processing Systems},
  33:7537--7547, 2020.

\bibitem{ucar2021subtab}
Talip Ucar, Ehsan Hajiramezanali, and Lindsay Edwards.
\newblock Subtab: Subsetting features of tabular data for self-supervised
  representation learning.
\newblock {\em Advances in Neural Information Processing Systems},
  34:18853--18865, 2021.

\bibitem{van2008visualizing}
Laurens Van~der Maaten and Geoffrey Hinton.
\newblock Visualizing data using t-sne.
\newblock {\em Journal of machine learning research}, 9(11), 2008.

\bibitem{wang2018tem}
Xiang Wang, Xiangnan He, Fuli Feng, Liqiang Nie, and Tat-Seng Chua.
\newblock Tem: Tree-enhanced embedding model for explainable recommendation.
\newblock In {\em Proceedings of the 2018 world wide web conference}, pages
  1543--1552, 2018.

\bibitem{yang2023uniaudio}
Dongchao Yang, Jinchuan Tian, Xu~Tan, Rongjie Huang, Songxiang Liu, Xuankai
  Chang, Jiatong Shi, Sheng Zhao, Jiang Bian, Xixin Wu, et~al.
\newblock Uniaudio: An audio foundation model toward universal audio
  generation.
\newblock {\em arXiv preprint arXiv:2310.00704}, 2023.

\bibitem{yoon2020vime}
Jinsung Yoon, Yao Zhang, James Jordon, and Mihaela van~der Schaar.
\newblock Vime: Extending the success of self-and semi-supervised learning to
  tabular domain.
\newblock {\em Advances in Neural Information Processing Systems},
  33:11033--11043, 2020.

\bibitem{zhang2023meta}
Yiyuan Zhang, Kaixiong Gong, Kaipeng Zhang, Hongsheng Li, Yu~Qiao, Wanli
  Ouyang, and Xiangyu Yue.
\newblock Meta-transformer: A unified framework for multimodal learning.
\newblock {\em arXiv preprint arXiv:2307.10802}, 2023.

\bibitem{zhu2021converting}
Yitan Zhu, Thomas Brettin, Fangfang Xia, Alexander Partin, Maulik Shukla,
  Hyunseung Yoo, Yvonne~A Evrard, James~H Doroshow, and Rick~L Stevens.
\newblock Converting tabular data into images for deep learning with
  convolutional neural networks.
\newblock {\em Scientific reports}, 11(1):11325, 2021.

\end{thebibliography}
\bibliographystyle{plain}

% --------------    
\newpage
\section*{Appendix}

\subsection*{More Results on partial-scale comparisons between NN Models}
\begin{figure}[h!]
\centering
\includegraphics[width=.6\linewidth]{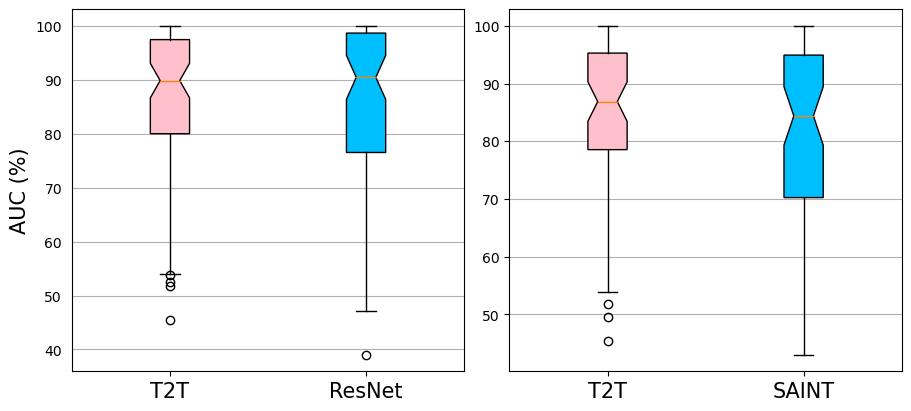}
\caption{Distribution of AUC (\%) for partial-scale comparison}
\label{fig:boxplot_partial_scale}
\end{figure}

We present the comparison of T2V, T2T, SAINT and ResNet on 59 intersected datasets in Table \ref{tab:rank_result_nn}. Similar to the observations reported in the partial-scale comparison, we find T2V outperforms SAINT but slightly underperforms ResNet. As showed in Figure \ref{fig:boxplot_partial_scale}, T2T does not generalize well on several datasets which limit its performance on average. 

\begin{table}[h]
    \centering
    \begin{tabular}{c|c c c c|c}
    \toprule 
    Algorithm & \multicolumn{4}{|c|}{Rank $\downarrow$} & AUC (\%) $\uparrow$ \\ 
    \midrule
     & min & max & mean & median & mean \\
    \midrule 
    ResNet & $1$ & $4$ & $2.15$ & $2$ & $84.87$ \\
    \midrule 
    T2T & $1$ & $4$ & $2.29$ & $2$ & $84.72$ \\
    \midrule
    T2V & $1$ & $4$ & $2.61$ & $3$ & $83.92$ \\
    \midrule
    SAINT & $1$ & $4$ & $3.01$ & $3$ & $81.46$ \\
    \bottomrule
    \end{tabular}
    \caption{Comparison between NN models on intersection datasets}
    \label{tab:rank_result_nn}
\end{table}

\begin{figure}[h!]
\centering
\includegraphics[width=.6\linewidth]{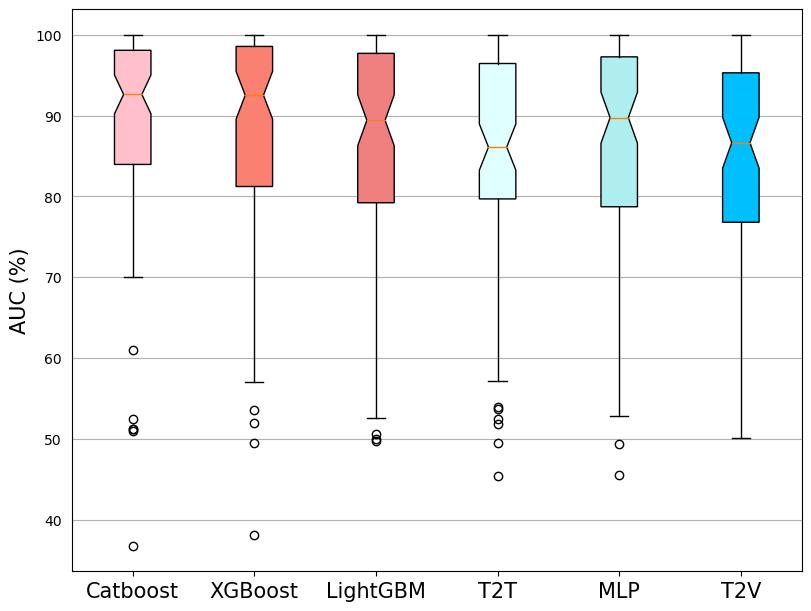}
\caption{Distribution of AUC (\%) for full-scale comparison}
\label{fig:boxplot_full_scale}
\end{figure}

\subsection*{Tree-to-Vector algorithms}
\begin{figure}[H]
\centering
\begin{subfigure}{.5\textwidth}
  \centering
  \includegraphics[width=.9\linewidth]{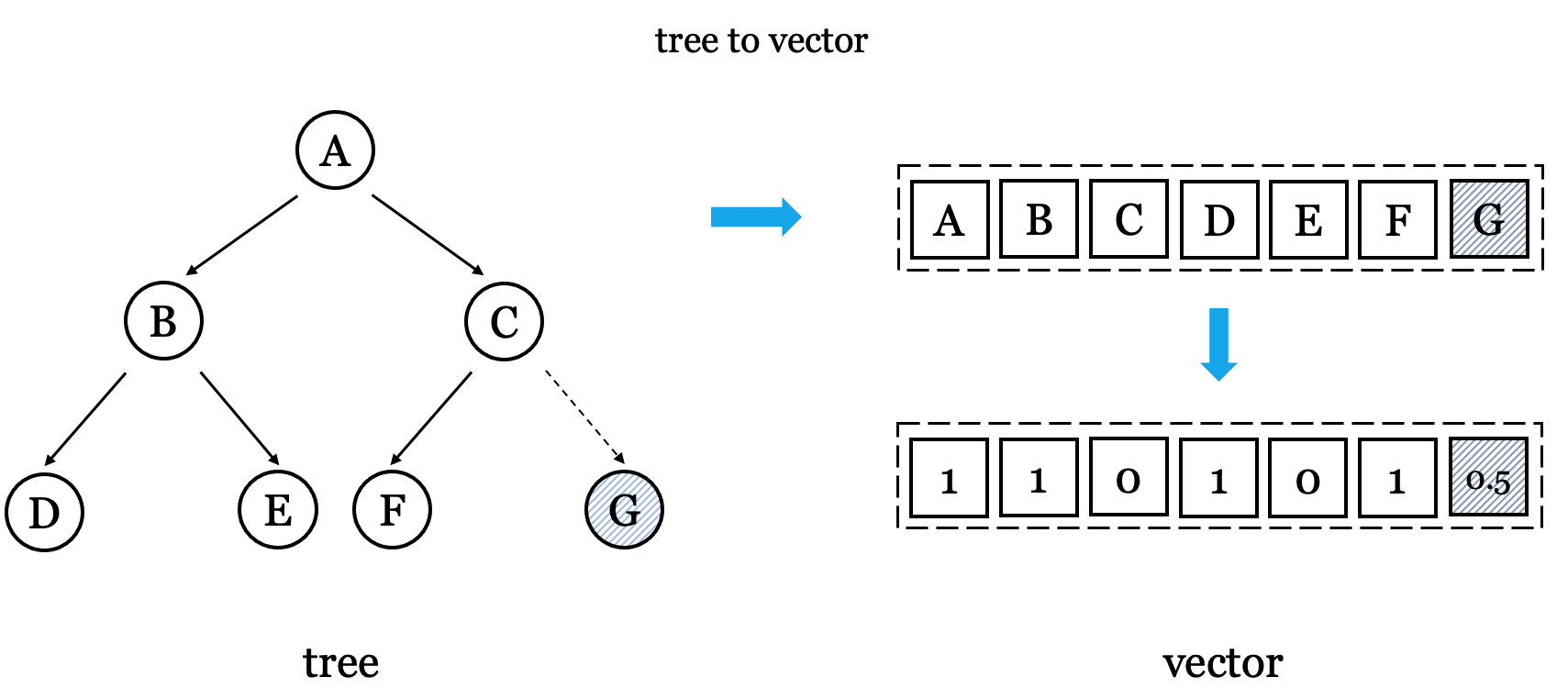}
  \caption{T2T: extract node. The nodes are traversed \\ in level order to maintain tree structure.}
  \label{fig:t2v_flowchart}
\end{subfigure}%
\begin{subfigure}{.5\textwidth}
  \centering
  \includegraphics[width=.9\linewidth]{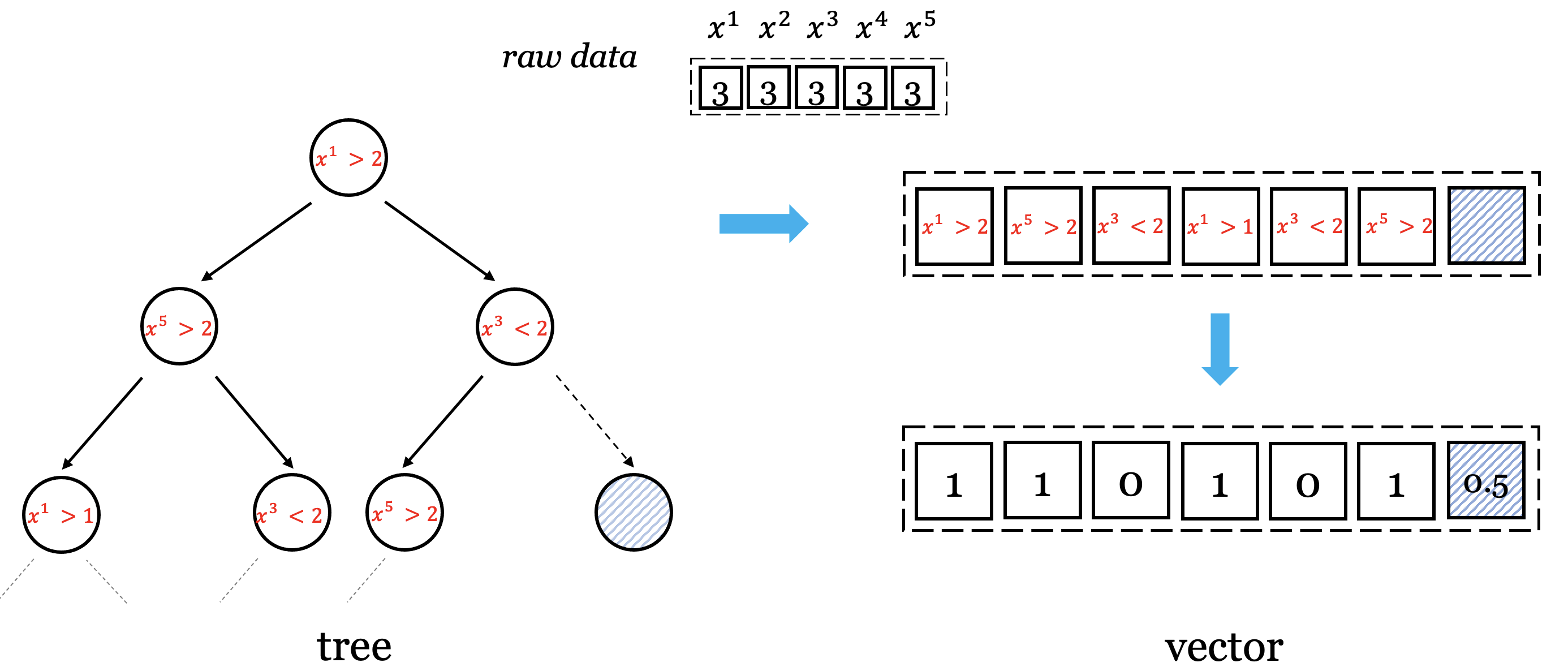}
  \caption{T2T: binary encode. A pseudo node G is added to make the tree complete and infilled with $0.5$ by default.}
  \label{fig:t2v_flowchart_example}
\end{subfigure}
\caption{An illustrative example of  T2T embedding generation}
\label{fig:t2v_example}
\end{figure}

We introduce T2V and T2T in Algorithm \ref{alg:t2v_I} and \ref{alg:t2v_II} respectively. For T2V, we set $\epsilon=4$, i.e., the thresholds are rounded with 4 digit of decimals. For T2T, we set $\tau = 0.5$ and $\eta = -1.0$, where the former is the default value to fill the complete tree and the later the default value to pad each token. The flowchart of T2V with an illustrative example is showed in Figure \ref{fig:t2v_example}.

\begin{algorithm}
\caption{Tree to Vector (T2V)}\label{alg:t2v_I}
\KwIn{xgb\_trees, $\epsilon$}
\KwOut{emb\_map} 
\kwInit{$\text{emb\_map} = \{\}$}
\For{$\mathrm{tree} \: \in \: \mathrm{xgb\_trees} \:$}{
    \For{$\mathrm{node} \: \in \: \mathrm{tree}  \:$}{
        $\{\text{var\_key, var\_val\}} = \text{node}$\;
        $\mathrm{var\_val.round(\epsilon)}$\;
        \If{$ \{\mathrm{var\_key, var\_val}\} \: \notin \: \mathrm{emb\_map}$}{
        $ \mathrm{emb\_map[var\_key].append(var\_val)} $; % \Comment*[r]{This is a comment}
        }
    }
}
\end{algorithm}

\begin{algorithm}
\caption{Tree to Tokens (T2T)}\label{alg:t2v_II}
\KwIn{xgb\_trees, $\tau, \: \eta$}
\KwOut{emb\_vec} 
\kwInit{$\text{vec\_len} = 0, \:\: \text{emb\_vec}=[]$}
\For{$\mathrm{tree} \: \in \: \mathrm{xgb\_trees} \:$}{
    $l = \text{tree.count\_node()}$ \; 
    $\text{vec\_len} = \max(\text{vec\_len}, \: l)$
}
\For{$\mathrm{tree} \: \in \: \mathrm{xgb\_trees} \:$}{
    $\mathrm{vec} = \mathrm{tree.to\_vec(\tau)}$\;
    $\mathrm{vec.pad(vec\_len, \: \eta)}$\; 
    $\mathrm{emb\_vec.append(vec)}$\; 
}
\end{algorithm}

\subsection*{OpenML Datasets}
task id: 7592, 9946, 49, 3797, 168911, 190410, 14951, 168912, 146606, 9977, 125920, 146607, 3903, 24, 3735, 3891, 3711, 9971, 167141, 27, 10089, 9965, 146820, 145984, 3485, 146065, 10101, 146047, 146819, 10093, 168338, 9952, 167125, 3731, 3561, 189354, 3917, 43, 3602, 4, 167211, 48, 3954, 9976, 9978, 3779, 3543, 219, 3953, 50, 9957, 168335, 3904, 3620, 3647, 3913, 14954, 146210, 29, 3896, 37, 3739, 145847, 189356, 39, 42, 3902, 3950, 3889, 3918, 145799, 3540, 31, 9910, 9984, 168337, 168868, 167120, 34539, 25, 15, 146206, 14952, 3748, 3686, 3, 54, 190408, 14965, 146818, 168908.

\end{document}